\pdfoutput=1
\documentclass[nonblindrev]{informs3_arXiv}

\OneAndAHalfSpacedXI 

\usepackage[draft]{hyperref}
\RequirePackage{fix-cm} 
\usepackage{xspace}
	\newcommand\ie{i.\,e.\xspace}
	\newcommand\eg{e.\,g.\xspace}

\usepackage{siunitx}
    \DeclareSIUnit\eur{\officialeuro}
    \DeclareSIUnit\M{M}
    \DeclareSIUnit\k{k}
	
	\usepackage[%
  sort&compress
]{cleveref}
  \crefname{chapter}{section}{sections}
	\Crefname{chapter}{Section}{Sections}

\usepackage{etoolbox}
\robustify\bfseries

\usepackage{url}

\usepackage{isomath}
\usepackage{bm}

\usepackage{colortbl}
\usepackage{tabularx}
\usepackage{booktabs}
\usepackage{collcell}
\usepackage{subfig}

\usepackage{float}
\usepackage{rotating}

\usepackage{array}
\newcolumntype{L}[1]{>{\raggedright\let\newline\\\arraybackslash\hspace{0pt}}p{#1}}
\newcolumntype{C}[1]{>{\centering\let\newline\\\arraybackslash\hspace{0pt}}p{#1}}
\newcolumntype{R}[1]{>{\raggedleft\let\newline\\\arraybackslash\hspace{0pt}}p{#1}}

\usepackage{csquotes}
\usepackage{amsmath}
\usepackage{amssymb}
\usepackage{pifont}

\usepackage[
    colorinlistoftodos,
    textsize=footnotesize,
        ]{todonotes}

\newcommand{\REWRITE}[1]{{\color{black}\hypersetup{citecolor=black}#1}}
\newcommand{\REVISION}[1]{{\color{black}\hypersetup{citecolor=black}#1}}        
\makeatletter
    \renewcommand{\fps@figure}{H}         
    \renewcommand{\fps@table}{H}         
\makeatother 

\usepackage{float}
\usepackage{rotating}

\makeatletter
\newcolumntype{B}[3]{>{\boldmath\DC@{#1}{#2}{#3}}c<{\DC@end}}
\makeatother

\usepackage{natbib}
 \bibpunct[, ]{(}{)}{,}{a}{}{,}%
 \def\bibfont{\small}%
\TheoremsNumberedThrough     

\EquationsNumberedThrough    

\begin{document}


 \RUNAUTHOR{Lutz, Pr\"ollochs, and Neumann}

\RUNTITLE{Review Length and Argumentation Changes}

\TITLE{The Longer the Better? The Interplay Between Review Length and Line of Argumentation in Online Consumer Reviews}

\ARTICLEAUTHORS{%
\AUTHOR{Bernhard Lutz}
\AFF{University of Freiburg, \EMAIL{bernhard.lutz@is.uni-freiburg.de}, \URL{}}
\AUTHOR{Nicolas Pr\"ollochs}
\AFF{University of Giessen, \EMAIL{nicolas.proellochs@wi.jlug.de} \URL{}}
\AUTHOR{Dirk Neumann}
\AFF{University of Freiburg, \EMAIL{dirk.neumann@is.uni-freiburg.de} \URL{}}
} 

\ABSTRACT{Review helpfulness serves as focal point in understanding customers' purchase decision-making process on online retailer platforms. An overwhelming majority of previous works find longer reviews to be more helpful than short reviews. In this paper, we propose that longer reviews should not be assumed to be uniformly more helpful; instead, we argue that the effect depends on the line of argumentation in the review text. To test this idea, we use a large dataset of customer reviews from Amazon in combination with a state-of-the-art approach from natural language processing that allows us to study argumentation lines at sentence level. Our empirical analysis suggests that the frequency of argumentation changes moderates the effect of review length on helpfulness. Altogether, we disprove the prevailing narrative that longer reviews are uniformly perceived as more helpful. Our findings allow retailer platforms to improve their customer feedback systems and to feature more useful product reviews.
}


\HISTORY{}

\maketitle

\section*{Introduction}
\label{sec:introduction}


\REWRITE{Customer reviews on online retailer platforms provide a valuable information source for customers before making purchase decisions \citep{Yin.2016}. An interesting feature of modern customer feedback systems is that they also allow to rate the perceived helpfulness of a product review \citep{Mudambi.2010}. 
\REWRITE{Previous studies have demonstrated that more helpful customer reviews have a greater influence on retail sales} \citep{Dhanasobhon.2007}.} Research on review helpfulness has received increasing attention lately, mainly because it serves as focal point for analyzing purchase decision-making \citep{Mudambi.2010}. For example, previous works have found that the review rating is an important determinant of review helpfulness \citep[\eg][]{Pavlou.2006b}. \REWRITE{In addition to meta data, online customer reviews typically contain review texts detailing customer opinions or user experiences \citep{Zimmermann.2018}.} 
\REWRITE{An overwhelming majority of previous works identify the length of the review text, \eg the number of sentences,} as a key explanatory variable and unanimously find longer reviews to be more helpful than short reviews \citep[\eg][]{Mudambi.2010,Pan.2011,Yin.2016}. 
 A plausible explanation is that longer reviews tend to be more diagnostic as they can provide more arguments about product quality and previous experiences \citep{Korfiatis.2012}. 


In this paper, however, we propose that longer reviews should not be assumed to be uniformly more helpful. Instead, we argue that the effect depends on the line of argumentation in the review text. Specifically, we suggest that frequent changes between positive and negative arguments require greater cognitive effort and may result in situations of information overload \citep{Jacoby.1977}. As a result, it may become difficult for customers to comprehend the review; and thus the review is unlikely to facilitate the purchase decision-making process. For example, it is an intriguing notion to expect long reviews, jumping excessively between positive and negative arguments, to be not particularly for customers. In contrast, a review providing a clear-cut, one-sided opinion or a support-then-refute order of positive and negative arguments may be easier to comprehend and also more persuasive. 
Therefore, we expect a higher frequency of argumentation changes in reviews to decrease perceived helpfulness. Moreover, given increased complexity and consumers' limited cognitive capacities, the (positive) effect of review length on perceived review helpfulness should be moderated by the frequency of argumentation changes in the review text. 


\REWRITE{To test these ideas, this paper examines the effects of review length and argumentation changes on review helpfulness. For this purpose, we use a large dataset of customer reviews from Amazon together with a state-of-the-art approach from natural language processing that allows us to study the line of argumentation on the basis of individual sentences. Given only the review label, the method uses distributed text representations in combination with multi-instance learning to infer sentence polarity labels. Specifically, our model learns to assign similar sentences in reviews to the same polarity label, whereas an opposite polarity is assigned to differing sentences. 
} 
The order in which sentences with positive and negative polarity appear then allows us to detect argumentation changes. Concordant with our propositions, our analyses suggest that the frequency of argumentation changes moderates the effect of review length on helpfulness. 


\REWRITE{Our findings have important implications for Information Systems research and practice}: we challenge the prevalent narrative in IS research that longer reviews are perceived as more helpful in general. To the best of our knowledge, our paper is the first study demonstrating that argumentation patterns and review length are closely intertwined. From a practical perspective, our findings can directly assist retailers in presenting more helpful product reviews and optimizing their customer feedback systems. 



\section*{Research Hypotheses}
\label{sec:background}



We now derive our research hypotheses, all of which are based on the notion that seeking helpful pre-purchase information plays an important role in consumers' decision-making processes \citep{Engel.1982}. The goal of this information search is to reduce risk and uncertainty in order to make better purchase decisions \citep{Murray.1991}. 



A product review usually consists of a star rating and a textual description \citep{Willemsen.2011}. The review text is commonly used to describe the product quality and previous experiences with the product \citep{Zimmermann.2018}. Longer review texts are likely to contain more information \citep{Mudambi.2010}. \citet{Tversky.1974} find \REWRITE{that decision-makers are more confident when there are more justifications in favor of a decision. It has also been shown that managers' arguments are more persuasive if they provide more information in support of the advocated position \citep{Schwenk.1986}. There are multiple factors contributing to this preference for diagnostic information. For example, a consumer may be inclined to purchase a product, but he/she has not yet made the necessary cognitive effort of identifying pros and cons of this product \citep{Mudambi.2010}.} In this scenario, a detailed review that provides a wide range of convincing arguments is likely to help the consumer make the purchase decision. Furthermore, the length of a review may reflect the reviewer's expertise. The more effort the reviewer puts into writing the review, the more likely it is that he/she will provide high quality information that aids others in making their purchase decisions \citep{Pan.2011}. Longer and more detailed reviews are also harder to fabricate, as a reviewer must have a certain degree of knowledge and experience to accurately describe different aspects of a product \citep{Jensen.2013}. Hence, it is reasonable to assume that longer reviews contain more elaborate arguments presented by better-informed reviewers that are more helpful to other customers. A positive effect of the length of a review on helpfulness has been suggested by a vast number of previous works. Our first hypothesis thus simply tests this link as discussed in the existing literature:

\vspace{.4em}
\emph{\textbf{Hypothesis~1 (H1).}} \textit{Longer consumer reviews are perceived as more helpful.}
\vspace{.4em}



A particularly relevant aspect of a review is the extent to which it is written in favor of or against the product. Reviews can be one-sided, \ie, arguing strictly for or against a product, or two-sided, enumerating pros and cons of a product. 
Existing literature has found that two-sided reviews are perceived as more credible \citep{Jensen.2013} and more helpful \citep[\eg][]{Lutz.2018}.
Yet \citet{Crowley.1994} note that the persuasiveness of two-sided argumentation is likely to depend on the mixture of positive and negative information. 
In a similar vein, \citet{Jackson.1987} argue that a two-sided message can be structured in three ways: (i) by starting with supporting arguments followed by opposing arguments, (ii) by starting with opposing arguments and then providing supportive arguments, or (iii) by interweaving supportive and opposing arguments. Hence, we expect that a relevant feature of two-sided reviews is the rate of argumentation changes, \ie how often the reviewer changes the line of argumentation from positive to negative and vice versa. 
\citet{Jackson.1987} find that a \enquote{support-then-refute order} is more persuasive than providing supporting and opposing arguments in an alternating manner. Providing arguments in an alternating manner also increases information entropy, \ie messages are not sufficiently organized as to be easily recognized as significant \citep{Hiltz.1985}. Altogether, we expect a higher rate of argumentation changes to present a less organized structure, which may make the review less helpful. 

\vspace{.4em}
\emph{\textbf{Hypothesis~2 (H2).}} \textit{A higher rate of argumentation changes decreases perceived review helpfulness.}
\vspace{.4em}



Following the above reasoning, an important question is whether review length and the rate of argumentation changes exhibit isolated effects on review helpfulness or rather depend on each other. Most consumer reviews are very one-sided in favor of or against a particular product \citep{Jensen.2013}. Strictly one-sided reviews do not change their line of argumentation from positive to negative or vice versa. 
Since a higher number of arguments in favor of a position makes a message more persuasive \citep[\eg][]{OKeefe.1998}, we expect longer reviews to be more helpful in situations in which the line of argumentation does not change between positive and negative arguments. 
In contrast, two-sided reviews enumerating pros and cons of a product change their argumentation at least once. We expect that processing a review with a high rate of argumentation changes requires greater cognitive effort than processing a review in which arguments are provided in clearly separated parts. A vast number of previous studies found that consumers' cognitive capacities are limited \citep[\eg][]{Bettman.1979}. Information overload theory suggests that consumers can process a certain amount and complexity of information, and that information which exceeds these capacities leads to poorer purchase decisions \citep{Jacoby.1977}. 
Hence, we expect that frequent changes between positive and negative arguments in long reviews can make it more difficult for customers to comprehend the review, thus moderating the positive effect of review length on helpfulness. 

\vspace{.4em}
\emph{\textbf{Hypothesis~3 (H3).}} \textit{The (positive) effect of review length on perceived review helpfulness is moderated by the rate of argumentation changes in the review text.} 

\section*{Dataset and Methodology}
\label{sec:Dataset}

This section presents our dataset. Subsequently, we make use of state-of-the-art methods from natural language processing for sentence-level polarity classification of texts. The order in which positive and negative sentences appear then allows us to determine argumentation changes in reviews.

\subsection*{Dataset}


\REWRITE{
To test our hypotheses, we use a large dataset of Amazon consumer reviews \citep{He.2016}. Compared to alternative review sources, this dataset exhibits several favorable characteristics. For example, the reviews are verified by Amazon and it is ensured that reviewers have actually purchased the product.} The Amazon platform also features a high number of retailer-hosted reviews per product due to a particularly active user base \citep{Gu.2012}. In addition, Amazon reviews are the prevalent choice in the related literature when studying review helpfulness \citep[see \eg][]{Gu.2012, Mudambi.2010}. Our dataset\footnote{We use the Amazon \emph{5-core} dataset available from \url{http://jmcauley.ucsd.edu/data/amazon/}. To account for possible imbalances, and to mitigate the effects of spammers, we focus on a review dataset which contains at most five reviews per reviewer. Moreover, we restrict our analysis to reviews that were created after 2010 and for which the helpfulness has been assessed at least once by other customers.} contains product reviews, ratings, and reviewer meta data for different product categories. In order to reduce our dataset to a reasonable size, we follow previous research \citep[\eg][]{Mudambi.2010, Ghose.2011} by restricting our analysis to a subset of product categories. We include all reviews from low-involvement products listed in the categories \emph{Groceries}, \emph{Music CDs}, and \emph{Videos} \citep{Kannan.2001}. These products feature a lower perceived risk of poor purchase decisions due to a lower price and lesser durability \citep{Gu.2012}. In addition, we include high-involvement product reviews listed in the categories \emph{Cell phones}, \emph{Digital cameras}, and \emph{Office electronics}. These products feature a higher price and greater durability, and hence a higher perceived risk \citep{Gu.2012}. 


Our complete dataset contains 51,837 Amazon customer reviews for 4647 low-involvement products and 2335 high-involvement products. Each review includes the following information: (i) the star rating assigned to the product (ranging between 1-5), (ii) the number of helpful and the number of unhelpful votes for the review, (iii) the review post date. 
Our reviews received between 0 and 4531 helpful votes, with a mean of 8.37. The mean star rating is 4.23. In addition, the corpus contains a textual description (the review text), which undergoes several preprocessing steps. \REVISION{First, we use the Stanford CoreNLP sentence-splitting tool \citep{Manning.2014} to split the review texts into sentences.} The length varies between one and 384 sentences, with a mean of 10.9 sentences. \REWRITE{Second, we use \emph{doc2vec} \citep{Le.2014} to create numerical representations of all sentences. This allows us to overcome some of the disadvantages of bag-of-words approach \citep[\eg][]{Prollochs.2016b,Prollochs.2019}, such as missing context \citep{Prollochs.2018}. The \emph{doc2vec} library uses a deep learning model to create numeric feature representations of text, which capture semantic information.} \REVISION{We use the hyperparameter settings as recommended by \citet{Lau.2016} and use the pre-trained word vectors from the Google News dataset to initialize the word vectors of the \emph{doc2vec} model \citep{Lutz.2019}.\footnote{\REVISION{The pretrained Google News dataset is a common choice when generating vector representations of Amazon reviews \citep[\eg][]{Kim.2015} and has several advantages \citep{Lau.2016,Kim.2015}: (1) tuning vector representations to a given dataset requires a large amount of training data; (2) the results are particularly robust and more reproducible.}}}

\subsection*{Sentence-Level Polarity Classification}

\REWRITE{The learning problem is a multi-instance learning task \citep{Dietterich.1997, Kotzias.2015}, in which we have to predict the polarity labels for all sentences in a set of reviews. Let $R$ denote the set of reviews, $K$ the number of reviews, $N$ the number of sentences, and $X = \{\boldsymbol{x}_i\}, i=1\dots N$ the set of all sentences. Each review $R_k=(\mathcal{G}_k, l_k)$ is represented by a multiset of sentences $\mathcal{G}_k \subseteq X$ with label $l_k$, which equals 1 for positive reviews and 0 for negative reviews. Given only the labels of the reviews, we then aim to learn a classifier $y_{\boldsymbol{\theta}}$ with hyperparameters $\boldsymbol{\theta}$ to predict individual sentence polarity labels $y_{\boldsymbol{\theta}}(\boldsymbol{x}_i)$.

Our multi-instance learning problem can be solved by optimizing a tailored loss function $L(\boldsymbol{\theta})$. The loss function consists of two components: first, a term punishing different labels for similar sentences. Second, a term punishing misclassifications at the document (review) level. Formally,
\small
\begin{align}
L(\boldsymbol{\theta}) &= \frac{1}{N^2} \sum\limits_{i=1}^N \sum\limits_{j=1}^N \mathcal{S}(\boldsymbol{x}_i,\boldsymbol{x}_j) (y_{\boldsymbol{\theta}}(\boldsymbol{x}_i) - y_{\boldsymbol{\theta}}(\boldsymbol{x}_j))^2 + \frac{\lambda}{K} \sum\limits_{k=1}^K (A(R_k,\boldsymbol{\theta}) - l_k)^2,  \label{eq:costgeneral}
\end{align}
\normalsize
where $\lambda$ is a hyperparameter that scales the prediction error at document~(review) level. The loss function is then minimized with respect to the classifier parameters $\boldsymbol{\theta}$. In \Cref{eq:costgeneral}, $\mathcal{S}(\boldsymbol{x}_i,\boldsymbol{x}_j)$ denotes a similarity measure between the representations of two sentences $\boldsymbol{x}_i$ and $\boldsymbol{x}_j$, $(y_{\boldsymbol{\theta}}(\boldsymbol{x}_i) - y_{\boldsymbol{\theta}}(\boldsymbol{x}_j))^2$ denotes the squared error between the predicted polarity labels for sentences $i$ and $j$, and $A(R_k,\boldsymbol{\theta})$ is the label that is predicted for review $R_k$. We adapt $L(\boldsymbol{\theta})$ to our problem of predicting sentence-level polarity labels by specifying the placeholders as follows: for measuring the similarity between two sentence representations, we use a radial basis function, \ie $\mathcal{S}(\boldsymbol{x}_i,\boldsymbol{x}_j) = e^{-||\boldsymbol{x}_i-\boldsymbol{x}_j||_2}$. While alternatives are possible, we use a logistic regression model for predicting $y_{\boldsymbol{\theta}}(\boldsymbol{x}_i)$ due to its interpretability and simplicity. Finally, $A(R_k,\boldsymbol{\theta})$ is defined as the average polarity label of all sentences in $\mathcal{G}_k$. The result is a specific loss function $L(\boldsymbol{\theta})$, which we minimize by $\boldsymbol{\theta}$. 
}
\subsection*{Determining Argumentation Changes in Reviews}


We use the aforementioned multi-instance learning approach to train a classifier for out-of-sample prediction of polarity label of sentences in reviews. For training the model, we use a \emph{disjunct} training dataset consisting of 5,000 positive and 5,000 negative reviews. The resulting classifier then allows us to predict a polarity label for each sentence in the dataset that is used in our later empirical analysis. As previously mentioned, we first transform each sentence in the corpus into its vector representation ($\boldsymbol{x}_i$). 
\REWRITE{Subsequently, the logistic regression model is used to calculate $y_{\boldsymbol{\theta}}(\boldsymbol{x}_i)$. If $y_{\boldsymbol{\theta}}(\boldsymbol{x}_i)$ is equal or greater than \SI{0.5}, sentence $i$ is assigned to a positive label, \ie $y_i=1$, and to a negative label otherwise. On an out-of-sample dataset of \num{1000} sentences (manually labeled), our approach achieves a classification accuracy of \SI{81.20}{\percent}. This can be regarded as sufficiently accurate in the context of our study.}


We then measure the rate of argumentation changes $RAC_k$ for review $R_k$ as follows. If the review consists of only a single sentence, then $RAC$ is defined as 0. For reviews that consist of at least two sentences, $RAC$ is defined as the number of argumentation changes divided by the length of the review in sentences minus 1,
\small
\begin{align}
RAC_k = \begin{cases}
0, & \, \text{if} \, |\mathcal{G}_k| = 1, \\
\frac{1}{|\mathcal{G}_k| - 1} \sum\limits_{i=2}^{|\mathcal{G}_k|} I(y_i \neq y_{i-1}), & \, \text{otherwise}, \label{eq:rac}
\end{cases}
\end{align}
\normalsize
where $|\mathcal{G}_k|$ denotes the number sentences of review $R_k$, and $I(cond)$ is an indicator function  which equals to 1, if $cond$ is true and 0 otherwise. 
Hence, $RAC$ is zero for one-sided reviews, and one for reviews in which the line of argumentation changes between each sentence. For example, a review consisting of five positive sentences followed by two negative sentences is mapped to the value $\frac{1}{7-1} = \frac{1}{6}$.

\section*{Preliminary Results}
\label{sec:results}


\subsection*{Empirical Model}


The target variable of our analysis is $RHVotes$. This variable denotes the number of users who voted \emph{Yes} in response to the question \enquote{Was this review helpful to you?}. The total number of users who responded to this question is denoted by $RVotes$. Following \citet{Pan.2011} and \citet{Yin.2016}, we model review helpfulness as a binomial variable with $RVotes$ trials. 


Concordant with previous works \citep[\eg][]{Mudambi.2010, Korfiatis.2012, Pan.2011, Yin.2016}, we incorporate the following variables to explain review helpfulness. First, we include the star rating of the review between 1 and 5 stars ($RStars$) and the average rating of the product ($PAvg$). Second, we control for the product type by adding a dummy that equals 1 for high-involvement products and 0 for low-involvement products ($PType$). Third, we control for multiple characteristics of the review text that may influence review helpfulness. Specifically, we calculate the fraction of cognitive and emotive words ($RCog$ and $REmo$) using LIWC 2015 
and control for readability using the Gunning-Fog index \citep{Gunning.1968} ($RRead$). The key explanatory variables for our research hypotheses are review length ($RLength$) and the rate of argumentation changes ($RAC$). To examine the interaction between review length and the rate of argumentation changes, we additionally incorporate an interaction term $RLength \times RAC$ into our model. Altogether, we model the number of helpful votes, $RHVotes$, as a binomial variable with probability parameter $\theta$ and $RVotes$ trials,
\vspace{-0.2cm}
\begin{flalign}
Logit(\theta) &= \beta_0 
	+ \beta_1 \,PAvg
	+ \beta_2 \,PType
	+ \beta_3 \,RAge
	+ \beta_4 \,RCog
	+ \beta_5 \,REmo
	+ \beta_6 \,RRead 	 
	+ \beta_7 \,RStars \nonumber  \\
	&+ \beta_8 \,RLength
	+ \beta_{9} \,RAC
	+ \beta_{10} \,RLength \times  RAC
	+ \alpha_{P} + \varepsilon , \label{eq:theta} \\ 
RHVotes & \sim Binomial[RVotes, \theta], \label{eq:binom}
\end{flalign}
with intercept $\beta_0$, a random intercept $\alpha_P$ for each product, and error term $\varepsilon$. 

\subsection*{Hypotheses Tests}


We estimate our model using \emph{mixed effects generalized linear models} and maximum likelihood estimation \citep{Wooldridge.2010}. The regression results are reported in \Cref{tbl:results}. To facilitate the interpretability of our findings, we z-standardize all variables so that we can compare the effects of regression coefficients on the dependent variable measured in standard deviations. Column (a) of \Cref{tbl:results} presents a baseline model that only includes the control variables from previous studies. We find that more recent reviews, higher star ratings, and reviews with a higher readability index are perceived as more helpful. In contrast, higher average ratings and higher shares of cognitive and emotive words have a negative effect. In addition, we find that high-involvement products tend to receive more helpful reviews.

\renewcommand{\arraystretch}{1.15}
\begin{table}[htp]
\TABLE
{Regression Linking Review Length and Argumentation Changes to Helpfulness \label{tbl:results}}
{
\footnotesize
\begin{tabular}{l SS SS SS}
\toprule
 & \multicolumn{4}{c}{\textbf{All Reviews}} & \multicolumn{2}{c}{\textbf{Review Subsets}}\\
 \cmidrule(lr){2-5} \cmidrule(lr){6-7}
 & \multicolumn{1}{c}{\textbf{(a)}} & \multicolumn{1}{c}{\textbf{(b)}} & \multicolumn{1}{c}{\textbf{(c)}}  & \multicolumn{1}{c}{\textbf{(d)}} & \multicolumn{1}{c}{\boldmath{$PType=0$}} & \multicolumn{1}{c}{\boldmath{$PType=1$}} \\
\midrule

$PAvg$                       & -0.078^{***} & -0.052^{***} & -0.052^{***} & -0.051^{***} & -0.001       & -0.099^{***} \\
                           & (0.012)      & (0.012)      & (0.012)      & (0.012)      & (0.015)      & (0.021)      \\ \addlinespace
$PType$                     & 0.479^{***}  & 0.349^{***}  & 0.349^{***}  & 0.360^{***}  &              &              \\
                           & (0.029)      & (0.028)      & (0.028)      & (0.028)      &              &              \\  \addlinespace
$RAge$                       & -0.304^{***} & -0.181^{***} & -0.181^{***} & -0.176^{***} & -0.082^{***} & -0.235^{***} \\
                           & (0.009)      & (0.009)      & (0.009)      & (0.009)      & (0.014)      & (0.011)      \\ \addlinespace
$RCog$                      & -0.022^{***} & -0.028^{***} & -0.029^{***} & -0.027^{***} & -0.060^{***} & -0.009       \\
                           & (0.006)      & (0.006)      & (0.006)      & (0.006)      & (0.009)      & (0.007)      \\ \addlinespace
$REmo$                       & -0.247^{***} & -0.131^{***} & -0.130^{***} & -0.126^{***} & -0.079^{***} & -0.162^{***} \\
                           & (0.006)      & (0.006)      & (0.006)      & (0.006)      & (0.009)      & (0.009)      \\ \addlinespace
$RRead$                      & 0.084^{***}  & 0.111^{***}  & 0.112^{***}  & 0.115^{***}  & 0.106^{***}  & 0.121^{***}  \\
                           & (0.005)      & (0.005)      & (0.005)      & (0.005)      & (0.009)      & (0.006)      \\  \addlinespace
$RStars$                     & 0.627^{***}  & 0.560^{***}  & 0.560^{***}  & 0.554^{***}  & 0.497^{***}  & 0.582^{***}  \\
                           & (0.004)      & (0.004)      & (0.004)      & (0.004)      & (0.007)      & (0.005)      \\ \addlinespace
$RLength$                    &              & 0.282^{***}  & 0.282^{***}  & 0.293^{***}  & 0.393^{***}  & 0.281^{***}  \\
                           &              & (0.003)      & (0.003)      & (0.003)      & (0.014)      & (0.003)      \\ \addlinespace
$RAC$                        &              &              & 0.010        & -0.036^{***} & -0.079^{***} & -0.013       \\
                           &              &              & (0.005)      & (0.006)      & (0.013)      & (0.007)      \\  \addlinespace
$RLength \times RAC$       &              &              &              & -0.169^{***} & -0.182^{***} & -0.173^{***} \\
                           &              &              &              & (0.008)      & (0.023)      & (0.008)      \\  \addlinespace
Intercept                  & 1.154^{***}  & 1.155^{***}  & 1.155^{***}  & 1.163^{***}  & 1.172^{***}  & 1.497^{***}  \\
                           & (0.019)      & (0.018)      & (0.018)      & (0.018)      & (0.019)      & (0.021)      \\  \addlinespace
\midrule
Observations  & {51,837}           & {51,837}    & {51,837}  & {51,837}           & {23,146}        & {28,691}          \\ 
Log-likelihood             & {$-$83,474.8}   & {$-$77,972.8}   & {$-$77,971.1}   & {$-$77,731.9}   & {$-$26,359.2}   & {$-$51,225.9}   \\
\bottomrule
\end{tabular}
}
{\hspace{-0.4cm} Stated: standardized coefficient and standardized error in parentheses. Significance:~$^{*}$p$<$0.05; $^{**}$p$<$0.01; $^{***}$p$<$0.001. Product-level effects are included. }
\end{table} 


To test H1, we additionally include the review length ($RLength$) in our model. The results are reported in Column (b) of \Cref{tbl:results}. We find that the coefficient of $RLength$ is statistically significant and positive ($\beta=0.282, p < 0.001$). This suggests that a one standard deviation increase in the length of the review text increases the probability of a helpful vote by $e^{0.282} -1 \approx 32.6\%$. The other coefficients in the model remain stable. Therefore, we find support for H1. For testing H2, we add the rate of argumentation changes ($RAC$) to our model. As shown in column (c) of \Cref{tbl:results}, $RAC$ is not statistically significant. Hence, H2 is rejected.

\begin{figure}[htp]
\FIGURE
{
    \includegraphics[width=8cm]{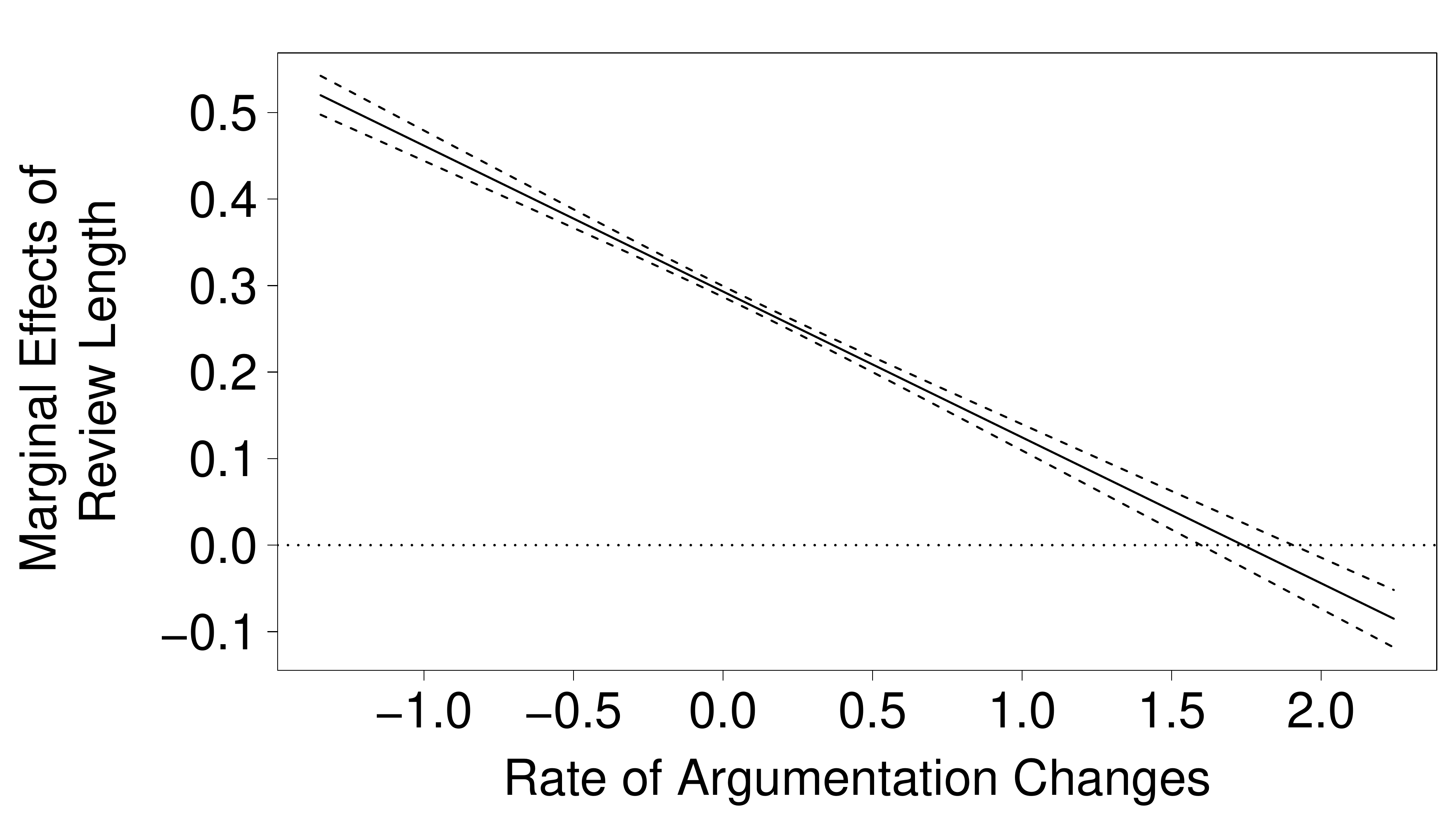}}
{\centering Standardized Marginal Effects of Review Length on Helpfulness \label{fig:interaction}}
{}
\end{figure}

Next, we add the interaction $RLength \times RAC$ to our model. This allows us to examine whether there is a significant interaction between review length and argumentation changes. Column (d) of \Cref{tbl:results} shows the results. The coefficient of the interaction term is negative and statistically significant ($\beta=-0.169, p < 0.001$), and the coefficient of $RAC$ became negative and significant ($\beta=-0.036, p < 0.001$).  
This suggests that the effects of review length and argumentation changes are interdependent. To shed light on the interaction, we plot the marginal effects of review length along with the \SI{95}{\percent} confidence intervals. \Cref{fig:interaction} shows that (i) the perceived helpfulness of long customer reviews is higher if the rate of argumentation changes is small, and (ii) longer reviews are perceived as less helpful if the rate of argumentation changes is very high. We thus find support for H3, which states that the positive effect of review length is moderated by the rate of argumentation changes.

Ultimately, we perform several checks and complementary analyses. First, we estimate two separate regressions for low- and high-involvement products. The results are shown in columns (e) and (f) of \Cref{tbl:results}. Concordant with our previous findings, we find that review length is moderated by the rate of argumentation changes. Interestingly, we further observe that the coefficient of $RAC$ is only significant for low-involvement products. A possible explanation is that customers prefer clear-cut opinions for low-involvement products as these products typically exhibit a relatively low amount of perceived risk. Second, we tested an alternative variant for measuring $RAC$ that additionally accounts for neutral sentences \citep{Ghose.2011}. This approach yields qualitatively identical results. Ultimately, we repeat our analysis using a mixed-effects tobit model as suggested by \citet{Mudambi.2010}. All regression estimates support our findings.

\section*{Discussion and Future Research}
\label{sec:discussion}



\REVISION{
This work makes several contributions to research on electronic commerce and online word-of-mouth. Most importantly, we disprove the prevailing narrative in previous research \citep[\eg][]{Mudambi.2010, Yin.2016} that longer reviews are uniformly perceived as more helpful. Instead, we propose that frequent changes between positive and negative arguments require greater cognitive effort, which can lead to information overload. This can make it less likely for customers to perceive longer reviews as helpful. Our work thereby extends the experimental study from \citet{Park.2008}, which indicates that information overload can occur at \emph{product level} such that consumers' involvement with a product is reduced if confronted with too many reviews. 
Our study provides evidence that information overload can also occur at the \emph{review level}. Specifically, to the best of our knowledge, our paper is the first study demonstrating that, given increased complexity and consumers' limited cognitive capacities, the (positive) effect of the length of a review on helpfulness is moderated by the frequency of argumentation changes in the review text.
}


In addition, our findings have important implications for practitioners in the field of electronic commerce. Retailers need to understand the determinants of review helpfulness in order to gain a better understanding of consumer information search behavior and purchase decision-making. Our findings and the proposed method for measuring the line of argumentation in reviews can help retailers to optimize their information systems towards a better shopping experience, \eg by improving the ranking of the most helpful reviews. The order in which reviews appear plays a crucial role, since most online platforms prominently display the most helpful positive and negative reviews, before presenting other reviews \citep{Yin.2016}. Our findings are also relevant for reviewers on retailer platforms, who can use our conclusions to write more helpful product reviews. Specifically, our study suggests that reviewers should avoid excessive alternation between positive and negative arguments, as this may make it more difficult to comprehend the review.


Overall, this work allows to better understand the effects of review length and argumentation changes on the helpfulness of consumer reviews. In future work, we will expand this study in three directions. First, we plan to study the interplay between review length and argumentation changes in the context of refutational and non-refutational reviews. Second, we will conduct further analysis to better understand potential differences regarding the role of argumentation changes for high-involvement and low-involvement products. Third, it is an intriguing notion to validate our findings with data from other recommendation platforms, such as hotel or restaurant reviews.

\bibliographystyle{misqNic}
\bibliography{bib/literature}

\end{document}